\documentclass[conference]{IEEEtran}
\IEEEoverridecommandlockouts
\usepackage{cite}
\usepackage{amsmath,amssymb,amsfonts}
\usepackage{algorithmic}
\usepackage{graphicx}
\usepackage{textcomp}
\usepackage{xcolor}
\def\BibTeX{{\rm B\kern-.05em{\sc i\kern-.025em b}\kern-.08em
    T\kern-.1667em\lower.7ex\hbox{E}\kern-.125emX}}

\usepackage{tabularx}
\usepackage{svg}
\usepackage{graphicx}
\usepackage{subfig}
\usepackage{bm}
\usepackage{multirow}%
\newcounter{magicrownumbers}
\newcommand\rownumber{\stepcounter{magicrownumbers}\arabic{magicrownumbers}}

\begin{document}

\newcommand{\fullname}{Metadata Enhanced Transformer}
\newcommand{\shortname}{SentiFormer}

%

\title{SentiFormer: Metadata Enhanced Transformer \\for Image Sentiment Analysis
}



\author{
\textit{Bin Feng$^{1,2}$, Shulan Ruan$^{3}$, Mingzheng Yang$^{1,2}$, Dongxuan Han$^{1,2}$, Huijie Liu$^{1,2}$, Kai Zhang$^{1,2}$, Qi Liu$^{1,2*}$\thanks{*Corresponding author. This research was partially supported by grants from the National Natural Science Foundation of China (Grant No. 62337001), the Key Technologies R\&D Program of Anhui Province (No. 202423k09020039), the Fundamental Research Funds for the Central Universities, and the National Key Laboratory of Human-Machine Hybrid Augmented Intelligence, Xi'an Jiaotong University (No. HMHAI-202410).}} \\
$^1$University of Science and Technology of China 
$^2$State Key Laboratory of Cognitive Intelligence\\
$^3$Shenzhen International Graduate School, Tsinghua University \\
fengbin@mail.ustc.edu.cn, slruan@sz.tsinghua.edu.cn, qiliuql@ustc.edu.cn
}

\maketitle

\newcommand{\fb}[1]{\textcolor{purple}{#1}}

\begin{abstract}
As more and more internet users post images online to express their daily emotions, image sentiment analysis has attracted increasing attention.
Recently, researchers generally tend to design different neural networks to extract visual features from images for sentiment analysis. 
Despite the significant progress, metadata, the data~(\textit{e.g.}, text descriptions and keyword tags) for describing the image, has not been sufficiently explored in this task.
In this paper, we propose a novel Metadata Enhanced Transformer for sentiment analysis (\shortname) 
to fuse multiple metadata and the corresponding image into a unified framework. 
Specifically, we first obtain multiple metadata of the image and unify the representations of diverse data. 
To adaptively learn the appropriate weights for each metadata, we then design an adaptive relevance learning module to highlight more effective information while suppressing weaker ones. 
Moreover, we further develop a cross-modal fusion module to fuse the adaptively learned representations and make the final prediction.
Extensive experiments on three publicly available datasets demonstrate the superiority and rationality of our proposed method.
\end{abstract}


\section{Introduction}

Image sentiment analysis aims to automatically predict the sentiment polarity expressed in images. It has a wide range of applications in many fields, such as opinion mining~\cite{al2024comprehensive} and recommendation system~\cite{zhan2023analyzing}. 
With the increasing popularity of social media platforms such as Flickr, Instagram and Twitter, more and more people tend to post images online to express their feelings and opinions of their daily lives. 
Therefore, this task has become a hot topic, and many significant efforts have been made in this field to help analyze image sentiment.

With the successful accomplishments of deep learning and computer vision, most recent studies focused on designing various networks to better extract image features for training sentiment polarity classifiers.
Early researchers~\cite{lu2012shape, zhao2014exploring, katsurai2016image} manually designed hand-crafted image features, such as color, shape, and texture, to explore the sentiment of images.
More recently, with the rapid popularity of CNNs, some studies~\cite{liang2021cross, yang2018weakly} focused on designing various CNN-based networks to extract deep features from images for improved sentiment analysis. 
Rao et al.~\cite{rao2020learning} and Koromilas et al.~\cite{koromilas2023mmatr}
used deep CNN networks for sentiment classification, and demonstrated the superior performance of deep features against hand-crafted features.
Rani et al.~\cite{rani2022efficient} combined CNN and LSTM to better integrate multi-level visual attributes for sentiment classification. 
With the popularity of attention mechanisms and transformer architectures, various neural network modules ~\cite{zhang2022image, 9102855, zhang2023learning, feng2024caption, ruan2021dae, truong2023concept} have been exploited by adaptively learning the attended image features.
Zhang et al.~\cite{zhang2022image} proposed stacked multi-head self-attention modules to explore the relationship between semantic regions and sentiment labels.
Truong et al.~\cite{truong2023concept} designed a concept-oriented transformer to capture the correlation between image features and specific concepts.
With the exploration of a unified architecture for large models, more recent works (\textit{e.g.}, CLIP~\cite{radford2021learning}, BLIP~\cite{li2022blip}, ImageBind~\cite{girdhar2023imagebind}) have been successively proposed.
Radford et al.~\cite{radford2021learning} utilized contrastive learning to embed images and text into a shared semantic space, enabling the learning of a universal vision-language representation.
Liu et al.~\cite{liu2024visual} utilized instruction-following data to fine-tune an end-to-end large model for general-purpose visual and language understanding.
These methods demonstrate their capacity to maintain comprehensive understanding across diverse data, resulting in remarkable performance in various tasks.


\begin{figure*}[t]
\centering
\includegraphics[width=.96\textwidth]{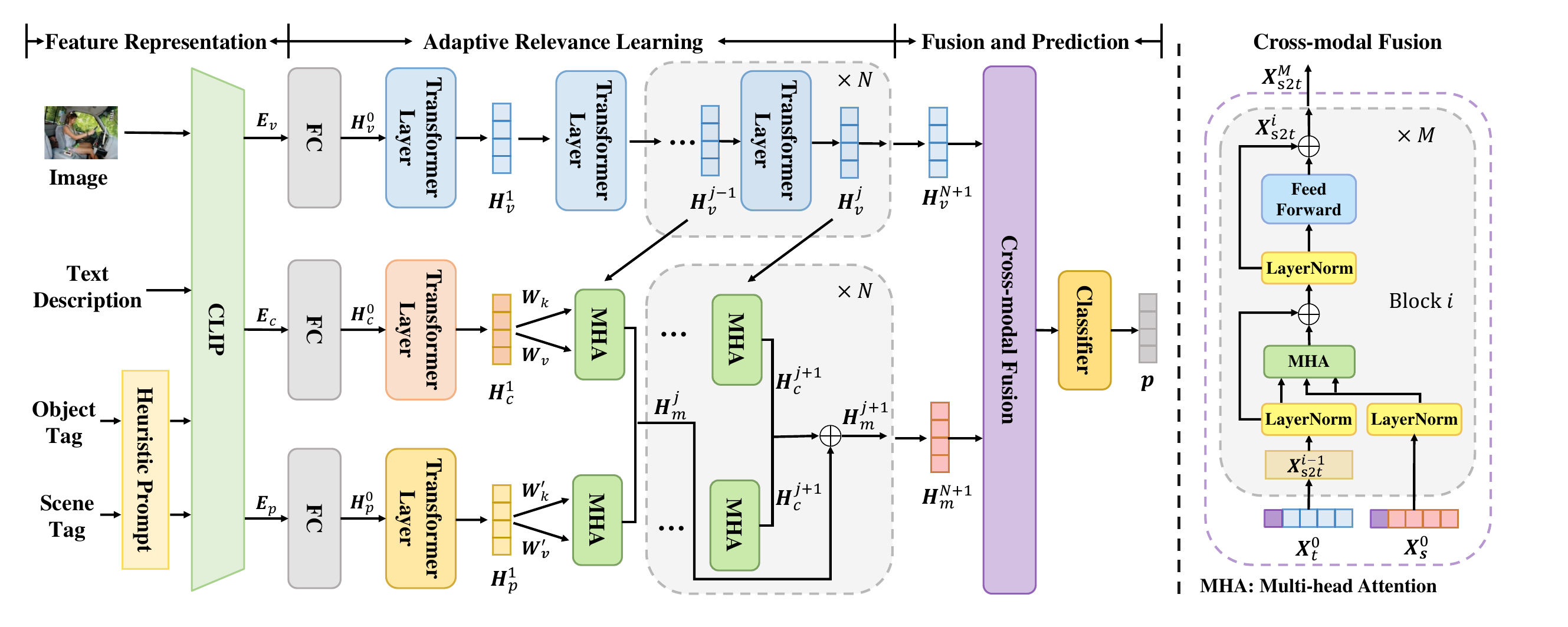}
\caption{The overall architecture of~\shortname.}
\label{fig:fig2}
\vspace{-6mm}
\end{figure*}

%
Despite the significant progress, metadata, the data~(\textit{e.g.}, text descriptions and keyword tags) for describing the image, has not been fully explored in this task, which is highly helpful for image sentiment understanding.
%
For example, considering a landscape photo with metadata of the scene tag, such as a beach or city, helps in understanding the image's context and sentiment. Knowing it was taken at a beach can suggest feelings of relaxation or joy, enhancing sentiment analysis accuracy.
This brings a new perspective to image sentiment analysis, which is the main focus of this paper: how to incorporate metadata into a unified framework and make comprehensive utilization of it for better knowledge reasoning and integration.

To this end, we propose a novel metadata enhanced transformer for image sentiment analysis (SentiFormer) to integrate multiple metadata and images into a unified framework.
Specifically, we first obtain multiple metadata corresponding to the image and use prompt learning for input alignment. 
Then, we employ CLIP to generate unified initial representations for both images and text. Next, we design an adaptive relevance learning module to highlight more effective information while suppressing weaker ones. Finally, we develop a cross-modal fusion module to integrate the adaptively learned representations and make the final prediction.
Moreover, we are the first to construct and publicly release metadata-enhanced image sentiment analysis datasets, which include text descriptions and keyword tags. 
All the data and code are publicly available at this link\footnote{https://github.com/MET4ISA/SentiFormer} for research purposes.
Extensive experiments on three benchmark datasets demonstrate the superiority and rationality of our proposed method.
%


\section{Model Structure}
\label{s:method}

As shown in Fig.~\ref{fig:fig2}, our proposed method consists of three modules:
the \textit{feature representation} module, the \textit{adaptive relevance learning} module, and the \textit{cross-modal fusion and prediction} module.

\subsection{Feature Representation}
\textbf{Metadata Representation:}
In social media, such as Flickr and Instagram, when users post an image to express their emotions, 
they usually attach metadata such as descriptions, tags, and so on.
Due to privacy reasons and copyright protection, some metadata are unprovided in many image sentiment analysis datasets. In order to get a full understanding of the impact of metadata on this task, for partial missing metadata, we adopt some methods to generate them with confidence.

Given an image $\bm{I}$, for the text description of the image, we utilize BLIP~\cite{li2022blip}, pre-trained on the COCO dataset, to generate textual caption $\bm{C}$.
Keyword tags typically include object tags and scene tags, which are highly helpful for image sentiment analysis.
In order to obtain objects contained in the image, we apply Faster R-CNN~\cite{ren2015faster} to obtain top-k object tags $\bm{T}_{obj}=\{\bm{obj}_{i}\}_{i=1}^{k}$ ranked by confidence score.
To obtain scene tag of the image, we employ Hiera~\cite{ryali2023hiera} to obtain the most likely one as $\bm{T}_{sce}$ among 365 categories in Place365~\cite{zhou2017places}.

For input alignment and better representation of multiple metadata,
we use prompt learning to represent keyword tags since they are word phrases rather than sentences. Specifically, we design a heuristic prompt $\bm{P} = $
\textit{`the scene or background of the image is $\bm{T}_{sce}$, and the image contains the following objects: $\bm{obj}_1$, $\bm{obj}_2$, ..., $\bm{obj}_k$'}.
%

\textbf{Image Representation:}
CLIP~\cite{radford2021learning} has powerful capabilities of image and text representation. Therefore, in this paper, we employ CLIP, which uses a ViT-B/32 transformer architecture as an image encoder and a masked self-attention transformer as a text encoder, to unify and align the representations of images and metadata.
After unified encoding, each output of $\bm{I}$, $\bm{C}$, and $\bm{P}$ above is a 512-dimensional vector, which can be represented as follows: 
\begin{equation}
    \bm{E}_v, \bm{E}_c, \bm{E}_p = \text{CLIP}(\bm{I, C, P}),
\end{equation}
where $\bm{E}_i \in \mathbb{R}^{d_e} , i \in \{ v, c, p \}$, and $d_e = 512$.

\subsection{Adaptive Relevance Learning}
\textbf{Unified Embedding:}
First, we pass $\bm{E}_v, \bm{E}_c, \bm{E}_p$ through a fully connected (FC) layer and project them into a $d_h$-dimensional space.
Then, we use three parallel transformer layers to unify the features of each part, respectively.
In order to better handle the $L$-classification task, we expand each feature to $L \times d_h$ dimensions in the transformer layer. Following Vision Transformer~\cite{dosovitskiy2020image}, the structure of each transformer layer includes layer normalization, multi-head self-attention, and a fully connected layer. In practice, we use 8-head attention and set the dimension of each head to 64. The process is formulated as follows:
\begin{eqnarray}
\begin{aligned}
    \bm{H}_i^0 &= \text{FC}(\bm{E}_i),\\
    \bm{H}_i^1 &= \text{Transformer}(({\bm{W}_i^0}{\bm{H}_i^0} + {\bm{b}_i^0})),
\end{aligned}
\end{eqnarray}
%
where $\bm{H}_i^0 \in \mathbb{R}^{1 \times d_h}$ and $\bm{H}_i^1 \in \mathbb{R}^{L \times d_h}$, $i \in \{ v, c, p \}$, and $\bm{W}_i^0 \in \mathbb{R}^{L \times 1} $, $\bm{b}_i^0 \in \mathbb{R}^{L \times d_h} $ are trainable parameters.

\textbf{Adaptive Learning:}
In image sentiment analysis, some metadata are more relevant to the image content, while others are irrelevant information or noise. In order to adaptively learn the appropriate weight for each metadata, we use the 
multi-head attention (MHA) mechanism
to highlight more effective information while suppressing the weaker.
Meanwhile, in order to capture the low-level to high-level features of the image, we use multiple cascaded transformer layers to learn a sequence of visual features. 
\begin{eqnarray}
\begin{aligned}
    \bm{H}_v^{j+1} &= \text{Transformer}(\bm{H}_v^{j}),
\end{aligned}
\end{eqnarray}
where $j = 1, ..., N$, and $N$ is the number of transformer layers.


In the interaction between image and metadata, we take this sequence of image features as query, and metadata embedding as key and value. 
Furthermore, in order to learn the joint representation of multiple metadata, we first randomly initialize a new token $\bm{H}_m^1 \in \mathbb{R}^{L \times d_h}$. Then, inspired by ResNet~\cite{he2016deep}, we introduce the residual block to gradually accumulate and update the learned metadata representation $\bm{H}_m^j$. 
The process is formulated as follows:
\begin{eqnarray}
\begin{aligned}
\text{MHA}(\bm{Q}, \bm{K}, \bm{V}) &= \textrm{Softmax}\left(\frac{\bm{Q}\bm{W}_Q\bm{W}_K^T\bm{K}^T}{\sqrt{d_k}}\right)\bm{V}\bm{W}_V, \\
\bm{H}_c^{j+1} &= \text{MHA}\left(\bm{H}_v^j, \bm{H}_c^1, \bm{H}_c^1\right),\\
\bm{H}_p^{j+1} &= \text{MHA}\left(\bm{H}_v^j, \bm{H}_p^1, \bm{H}_p^1\right),\\
\bm{H}_m^{j+1} &= \bm{H}_m^{j}+(\bm{H}_c^{j+1}+\bm{H}_p^{j+1})\bm{W}_O,
\end{aligned}
\end{eqnarray}
where $j = 1, ..., N$, and ${d_k}$ means the dimension of each head of the multi-head attention. $\bm{W}_Q$, $\bm{W}_K$, $\bm{W}_V$, and $\bm{W}_O$ are trainable parameters.


%
\subsection{Cross-modal Fusion and Prediction}
\textbf{Cross-modal Fusion:}
When predicting different sentiment labels, not all visual information and metadata make the same contribution to the final prediction. Therefore, we adopt a cross-modal transformer to tackle this problem.

We take $\bm{H}_v^{N+1}$ and $\bm{H}_m^{N+1}$ as input, and add an extra\_token $\bm{H}_e$ and position embedding.
$\bm{H}_e$ is a $d_h$-dimensional token, which is used to capture global information at the head of a sequence, and
position embedding is used to preserve sequential information of the input.
As shown in Fig.~\ref{fig:fig2}, the cross-modal transformer is a deep stacking of several cross-modal attention blocks, with a depth of $M$. In the cross-modal transformer, we first take $\bm{X}_t^0$ as the query and $\bm{X}_s^0$ as the key and value.
The cross-modal transformer can effectively capture and fuse information from diverse data to enrich feature representation, which can be formulated as follows.


First, the extra token is concatenated to the input sequence, followed by adding positional embeddings:
\begin{eqnarray}
\begin{aligned}
\bm{X}_s^0 &= \bm{H}_e \oplus \bm{H}_v^{N+1} + \bm{P}_s,\\
\bm{X}_t^0 &= \bm{H}_e \oplus \bm{H}_m^{N+1} + \bm{P}_t,
\end{aligned}
\end{eqnarray}
where \( \oplus \) denotes the concatenation operation along the sequence dimension, and \( \bm{H}_e \in \mathbb{R}^{ 1 \times d_h} \) represents the added token.
The positional embeddings are defined as $\bm{P}_s \in \mathbb{R}^{(L + 1) \times d_h}$ and$\quad \bm{P}_t \in \mathbb{R}^{(L + 1) \times d_h}$.

Next, in each transformer block, the multi-head attention computes the attention from the target to the source as follows:
\begin{eqnarray}
\begin{aligned}
[\bm{Q}_t, \bm{K}_s, \bm{V}_s] &= [\bm{X}_t \bm{W}_{Q'}, \bm{X}_s \bm{W}_{K'}, \bm{X}_s \bm{W}_{V'}],\\
\bm{X}_{s2t} &= \text{Softmax}\left(\frac{\bm{Q}_t \bm{K}_s^T}{\sqrt{d_s}}\right) \bm{V}_s,
\end{aligned}
\end{eqnarray}
where \( \bm{W}_{Q'}, \bm{W}_{K'}, \bm{W}_{V'}\) are trainable parameters, and \( d_s \) is the dimension of each attention head. 

Then, we obtain the final cross-modal fusion output $\bm{X}_{s2t}^M$ by repeating cross-modal transformer blocks $M$ times.

\textbf{Sentiment Prediction:}
For the image sentiment classification task, we select the first token of the sequence from the cross-modal fusion output. The token at the head of the sequence contains global information of the entire sequence. 
Then we construct a classification head to make the final prediction, which consists of a linear layer and a softmax layer.
\begin{eqnarray}
\begin{aligned}
\bm{p} &= \text{Softmax}(\bm{X}_{s2t}^M[:, 0] \bm{W}_f + \bm{b}_f),
\end{aligned}
\end{eqnarray}
where $\bm{W}_f \in \mathbb{R}^{d_h \times L}$ and $\bm{b}_f \in \mathbb{R}^{1 \times L}$ are trainable parameters.

For model learning, we employ the cross-entropy as the loss
function, which is calculated as follows:
\begin{eqnarray}
\begin{aligned}
L &= -\frac{1}{n} \sum_{i = 1}^{n} \boldsymbol{y}_{i} log P(\boldsymbol{p}_{i} \mid \boldsymbol{I}),
\end{aligned}
\end{eqnarray}
where $\boldsymbol{y}_i$ is the true answer label of the $i^{th}$ instance of the dataset, and $n$ represents the number of training instances.

\section{Experiment}
\label{s:experiment}

\subsection{Data Description}
We evaluate our method on FI~\cite{you2016building}, Twitter\_LDL~\cite{yang2017learning} and Artphoto~\cite{machajdik2010affective}.
FI is a public dataset built from Flickr and Instagram, with 23,308 images and eight emotion categories, namely \textit{amusement}, \textit{anger}, \textit{awe},  \textit{contentment}, \textit{disgust}, \textit{excitement},  \textit{fear}, and \textit{sadness}. 
Twitter\_LDL contains 10,045 images from Twitter, with the same eight categories as the FI dataset.
For these two datasets, they are randomly split into 80\%
training and 20\% testing set.
Artphoto contains 806 artistic photos from the DeviantArt website, which we use to further evaluate the zero-shot capability of our model.


%
\begin{table}[t]
\centering
\caption{Overall performance of different models on FI and Twitter\_LDL datasets.}
\label{tab:cap1}
{
\begin{tabular}{l|c|c|c|c}
\hline
\multirow{2}{*}{\textbf{Model}} & \multicolumn{2}{c|}{\textbf{FI}}  & \multicolumn{2}{c}{\textbf{Twitter\_LDL}} \\ \cline{2-5} 
  & \textbf{Accuracy} & \textbf{F1} & \textbf{Accuracy} & \textbf{F1}  \\ \hline
(\rownumber)~ResNet101~\cite{he2016deep} & 66.16\%& 65.56\%  & 62.02\% & 61.34\%  \\ 
(\rownumber)~CDA~\cite{han2023boosting} & 66.71\%& 65.37\%  & 64.14\% & 62.85\%  \\ 
(\rownumber)~CECCN~\cite{ruan2024color} & 67.96\%& 66.74\%  & 64.59\%& 64.72\% \\ 
(\rownumber)~EmoVIT~\cite{xie2024emovit} & 68.09\%& 67.45\%  & 63.12\% & 61.97\%  \\ 
(\rownumber)~ComLDL~\cite{zhang2022compound} & 68.83\%& 67.28\%  & 65.29\% & 63.12\%  \\ 
(\rownumber)~WSDEN~\cite{li2023weakly} & 69.78\%& 69.61\%  & 67.04\% & 65.49\% \\ 
(\rownumber)~ECWA~\cite{deng2021emotion} & 70.87\%& 69.08\%  & 67.81\% & 66.87\%  \\ 
(\rownumber)~EECon~\cite{yang2023exploiting} & 71.13\%& 68.34\%  & 64.27\%& 63.16\%  \\ 
(\rownumber)~MAM~\cite{zhang2024affective} & 71.44\%  & 70.83\% & 67.18\%  & 65.01\%\\ 
(\rownumber)~TGCA-PVT~\cite{chen2024tgca}   & 73.05\%  & 71.46\% & 69.87\%  & 68.32\% \\ 
(\rownumber)~OEAN~\cite{zhang2024object}   & 73.40\%  & 72.63\% & 70.52\%  & 69.47\% \\ \hline
(\rownumber)~\shortname  & \textbf{79.48\%} & \textbf{79.22\%} & \textbf{74.12\%} & \textbf{73.09\%} \\ \hline
\end{tabular}
}
\vspace{-6mm}
\end{table}

\subsection{Experiment Setting}
%
\textbf{Model Setting:}
For feature representation, we set $k=10$ to select object tags, and adopt clip-vit-base-patch32 as the pre-trained model for unified feature representation.
Moreover, we empirically set $(d_e, d_h, d_k, d_s) = (512, 128, 16, 64)$, and set the classification class $L$ to 8.

\textbf{Training Setting:}
To initialize the model, we set all weights such as $\boldsymbol{W}$ following the truncated normal distribution, and use AdamW optimizer with the learning rate of $1 \times 10^{-4}$.
Furthermore, we set the batch size to 32 and the epoch of the training process to 200.
During the implementation, we utilize \textit{PyTorch} to build our entire model.

\textbf{Evaluation Metrics:}
Following~\cite{zhang2024affective, chen2024tgca, zhang2024object}, we adopt \textit{accuracy} and \textit{F1} as our evaluation metrics to measure the performance of different methods for image sentiment analysis.

\subsection{Experiment Result}
We compare our model against several baselines, and the overall results are summarized in Table~\ref{tab:cap1}.
We observe that our model achieves the best performance in both accuracy and F1 metrics, significantly outperforming the previous models. 
This superior performance is mainly attributed to our effective utilization of metadata to enhance image sentiment analysis, as well as the exceptional capability of the unified sentiment transformer framework we developed. These results strongly demonstrate that our proposed method can bring encouraging performance for image sentiment analysis.

\setcounter{magicrownumbers}{0} 
\begin{table}[t]
\begin{center}
\caption{Ablation study of~\shortname~on FI dataset.} 
\label{tab:cap2}
\resizebox{.9\linewidth}{!}
{
\begin{tabular}{lcc}
  \hline
  \textbf{Model} & \textbf{Accuracy} & \textbf{F1} \\
  \hline
  (\rownumber)~Ours (w/o vision) & 65.72\% & 64.54\% \\
  (\rownumber)~Ours (w/o text description) & 74.05\% & 72.58\% \\
  (\rownumber)~Ours (w/o object tag) & 77.45\% & 76.84\% \\
  (\rownumber)~Ours (w/o scene tag) & 78.47\% & 78.21\% \\
  \hline
  (\rownumber)~Ours (w/o unified embedding) & 76.41\% & 76.23\% \\
  (\rownumber)~Ours (w/o adaptive learning) & 76.83\% & 76.56\% \\
  (\rownumber)~Ours (w/o cross-modal fusion) & 76.85\% & 76.49\% \\
  \hline
  (\rownumber)~Ours  & \textbf{79.48\%} & \textbf{79.22\%} \\
  \hline
\end{tabular}
}
\end{center}
\vspace{-5mm}
\end{table}

\begin{figure}[t]
\centering
\includegraphics[width=0.42\textwidth]{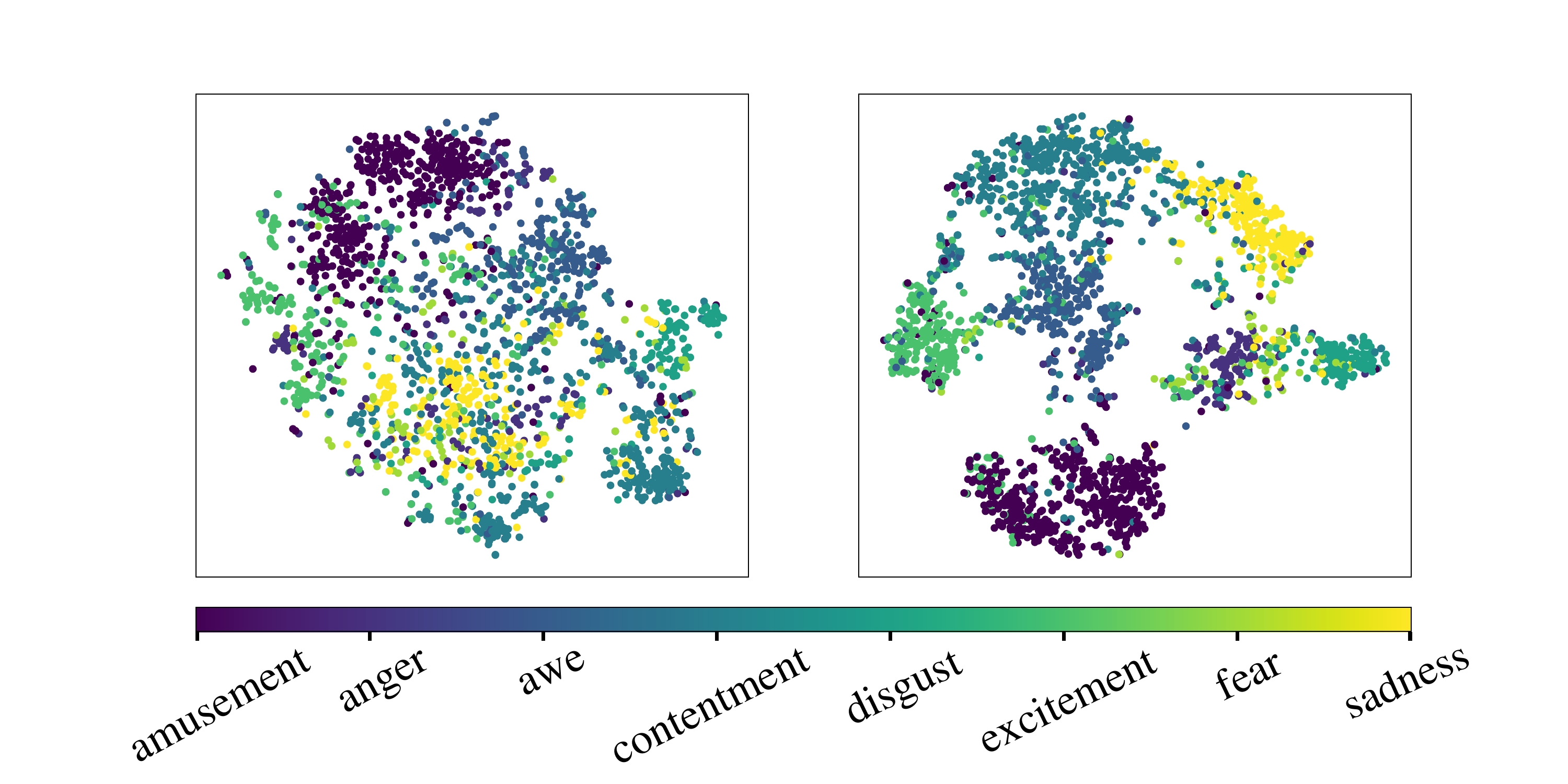}
\caption{Visualization of feature distribution on eight categories before (left) and after (right) model processing.}
\label{fig:visualization}
\vspace{-5mm}
\end{figure}

\subsection{Ablation Performance}
In this subsection, we conduct an ablation study to examine which component is really important for performance improvement. The results are reported in Table~\ref{tab:cap2}.

For information utilization, we observe a significant decline in model performance when visual features are removed. Additionally, the performance of \shortname~decreases when different metadata are removed separately, which means that text description, object tag, and scene tag are all critical for image sentiment analysis.
Recalling the model architecture, we separately remove transformer layers of the unified representation module, the adaptive learning module, and the cross-modal fusion module, replacing them with MLPs of the same parameter scale.
In this way, we can observe varying degrees of decline in model performance, indicating that these modules are indispensable for our model to achieve better performance.

\subsection{Visualization}
%




\begin{figure}[tbp]   
\vspace{-4mm}
  \centering            
  \subfloat[Depth of adaptive learning layers]   
  {
    \label{fig:subfig1}\includegraphics[width=0.22\textwidth]{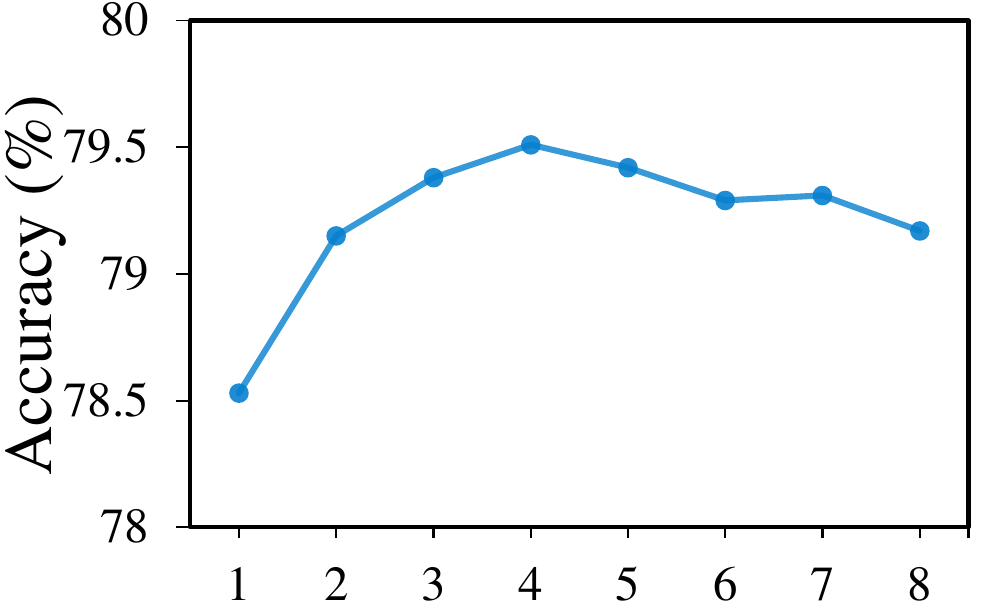}
  }
  \subfloat[Depth of fusion layers]
  {
    \label{fig:subfig2}\includegraphics[width=0.22\textwidth]{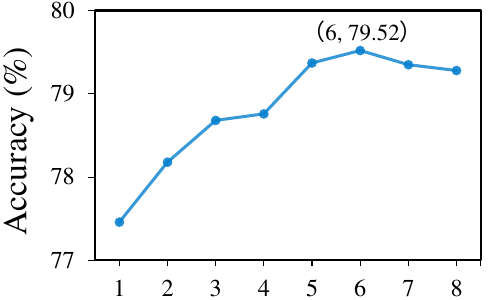}
  }
  \caption{Sensitivity study of \shortname~on different depth. }   
  \label{fig:fig_sensitivity}  
\vspace{-2mm}
\end{figure}



In Fig.~\ref{fig:visualization}, we use t-SNE~\cite{van2008visualizing} to reduce the dimension of data features for visualization.
The left figure shows metadata features before being processed by our model (\textit{i.e.}, embedded by CLIP), while the right shows the distribution of features after being processed by our model.
We can observe that after the model processing, data with the same label are closer to each other, while others are farther away.
Therefore, it shows that the model can effectively utilize the information contained in the metadata and use it to guide the classification process.

\subsection{Sensitivity Analysis}
In this subsection, we conduct a sensitivity analysis to figure out the effect of different depth settings of adaptive learning layers and fusion layers. 
Taking Fig.~\ref{fig:fig_sensitivity} (a) as an example, the model performance improves with increasing depth, reaching the best performance at a depth of 4.
When the depth continues to increase, the accuracy decreases to varying degrees.
Similar results can be observed in Fig.~\ref{fig:fig_sensitivity} (b).
Therefore, we set their depths to 4 and 6 respectively to achieve the best results.


\subsection{Zero-shot Capability}
%


\begin{table}[t]
\centering
\caption{Zero-shot capability of \shortname.}
\label{tab:cap3}
\resizebox{1\linewidth}{!}
{
\begin{tabular}{lc|lc}
\hline
\textbf{Model} & \textbf{Accuracy} & \textbf{Model} & \textbf{Accuracy} \\ \hline
(1)~WSCNet~\cite{2019WSCNet}  & 30.25\% & (5)~MAM~\cite{zhang2024affective} & 39.56\%  \\ \hline
(2)~AR~\cite{2018Visual} & 32.67\% & (6)~CECCN~\cite{ruan2024color} & 43.83\% \\ \hline
(3)~RA-DLNet~\cite{2020A} & 34.01\%  & (7)~EmoVIT~\cite{xie2024emovit} & 44.90\% \\ \hline
(4)~CDA~\cite{han2023boosting} & 38.64\% & (8)~Ours (Zero-shot) & 47.83\% \\ \hline
\end{tabular}
}
\vspace{-5mm}
\end{table}


To validate the model's generalization ability and robustness to other distributed datasets, we directly test the model trained on the FI dataset, without training on Artphoto. 
From Table~\ref{tab:cap3}, we can observe that compared with other models trained on Artphoto, we achieve competitive zero-shot performance, which shows that the model has good generalization ability in out-of-distribution tasks.

\section{Conclusion}
\label{s:conclusion}

In this paper, we introduced metadata into image sentiment analysis and proposed a unified metadata-enhanced sentiment transformer.
Specifically, we first obtained multiple metadata and unified their representations with images. Then, we further developed an adaptive relevance learning module and a cross-modal fusion module for better sentiment prediction.
Extensive experiments on three datasets demonstrated the superiority and rationality of our proposed method.

\bibliography{main}
\bibliographystyle{IEEEtran}

\end{document}